\documentclass[runningheads]{llncs}
\usepackage{graphicx}

\usepackage{tikz}
\usepackage{comment}
\usepackage{amsmath,amssymb} 
\usepackage{color}
\usepackage{dsfont}
\usepackage{multirow}
\usepackage{multicol}
\usepackage[font={normalsize}]{subfig}
\usepackage[ruled,noalgohanging]{algorithm2e}
\usepackage{wrapfig}
\usepackage{graphicx}

\usepackage{ctable}
\definecolor{myred}{RGB}{196, 0, 0}
\definecolor{myblue}{RGB}{70, 114, 196}
\definecolor{mygreen}{RGB}{0, 128, 64}

\usepackage{pifont}
\newcommand{\cmark}{\ding{51}}%
\newcommand{\xmark}{\ding{55}}%

\DeclareMathOperator*{\argmin}{arg\,min}

\usepackage[accsupp]{axessibility}  

\begin{document}
\pagestyle{headings}
\mainmatter
\def\ECCVSubNumber{4074}  

\title{Negative Samples are at Large: Leveraging Hard-distance Elastic Loss for Re-identification} 

\titlerunning{ECCV-22 submission ID \ECCVSubNumber} 
\authorrunning{ECCV-22 submission ID \ECCVSubNumber} 
\author{Anonymous ECCV submission}
\institute{Paper ID \ECCVSubNumber}

\titlerunning{Leveraging Hard-distance Elastic Loss for Re-identification}
%
\author{Hyungtae Lee$^{1}$~~~~~~~~~~
Sungmin Eum$^{1,2}$~~~~~~~~~~
Heesung Kwon$^{1}$}
\authorrunning{H. Lee et al.}
%
\institute{$^{1}$DEVCOM Army Research Laboratory ~~~~~~~~~~ $^{2}$Booz Allen Hamilton}
\maketitle

\begin{abstract}
We present a Momentum Re-identification (MoReID) framework that can leverage a very large number of negative samples in training for general re-identification task. The design of this framework is inspired by Momentum Contrast (MoCo), which uses a dictionary to store current and past batches to build a large set of encoded samples. As we find it less effective to use past positive samples which may be highly inconsistent to the encoded feature property formed with the current positive samples, MoReID is designed to use only a large number of negative samples stored in the dictionary. However, if we train the model using the widely used Triplet loss that uses only one sample to represent a set of positive/negative samples, it is hard to effectively leverage the enlarged set of negative samples acquired by the MoReID framework. To maximize the advantage of using the scaled-up negative sample set, we newly introduce Hard-distance Elastic loss (HE loss), which is capable of using more than one hard sample to represent a large number of samples. Our experiments demonstrate that a large number of negative samples provided by MoReID framework can be utilized at full capacity only with the HE loss, achieving the state-of-the-art accuracy on three re-ID benchmarks, VeRi-776, Market-1501, and VeRi-Wild.

\end{abstract}

\section{Introduction}
\label{sec:intro}

Re-identification (re-ID) is the task of finding instances (e.g., person, vehicle) with the same identity in images taken from different viewpoints in different locations. In general, a re-ID model needs the ability that, given a query, well separates instances with the same ID (positive samples) from instances with different IDs (negative samples). This ability can be acquired by training the model so that the positive samples are placed closer to the query while the negative samples are placed in a distant location in the learned feature space.

\begin{figure}[t]
    \captionsetup{font=small}
    \centering
    \includegraphics[trim=40mm 25mm 5mm 15mm,clip,width=0.8\linewidth]{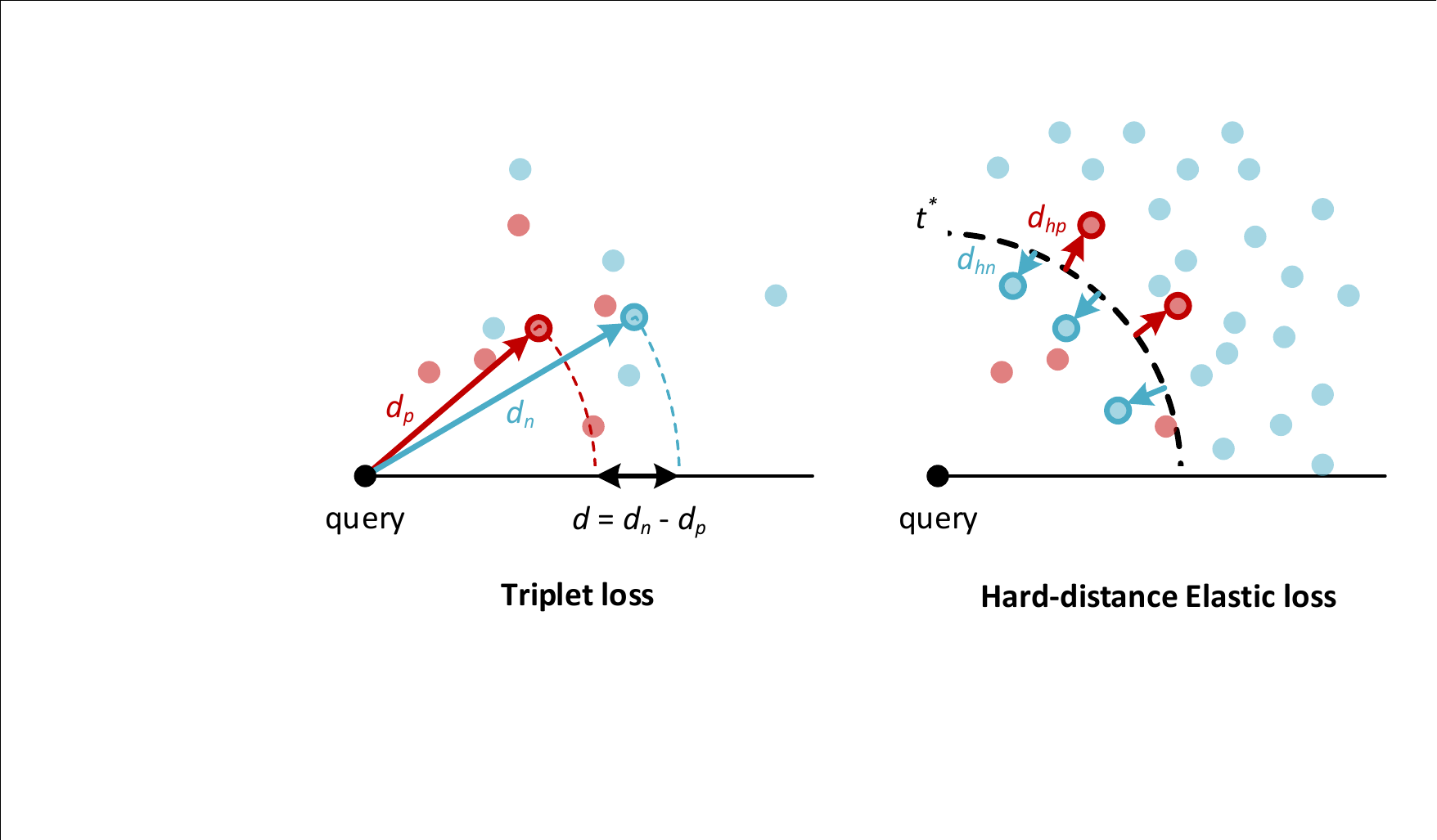}
    \vspace{-0.4cm}
    \caption{{\bf Hard-distance Elastic Loss vs. Triplet loss.} In Triplet loss, the positive and negative samples are each represented by a single point. HE loss, on the other hand, can use more than one hard samples to represent positive and negative samples, respectively. Knowing that the optimally determined boundary $t^*$ divides the feature space into positive and negative sides, a positive sample is a called a hard sample when it lies on the negative side, and vice versa. Representative samples are plotted in bold. Multiple points are better than a single point when representing a large number of negative samples, as shown in the figure on the right. More analysis of HE loss versus Triplet loss is in Section~\ref{ssec:loss_comparison}.}
    \label{fig:HE_Loss}
\end{figure}

When separating the positives from the negatives with respect to a given query, such as in the re-ID, having a sufficient number of samples in each set often helps in generating a more precise boundary between the two sets. However, the mini-batch optimization which is widely used in training a network cannot afford to use more than a certain number of samples in each training iteration. Recently, Momentum Contrast (MoCo)~\cite{KHeCVPR2020} provided a structural way to effectively provide a large number of samples for training. It uses a dictionary that collects several preceding batches in the form of encoded features. To fully take advantage of using a large number of samples when generating the precise boundary, features of these samples are desired to be encoded in a consistent feature space. To maintain this consistency as training progresses, MoCo uses an additional slow-progressing encoder that only processes samples to be stored in the dictionary (these samples are called ``keys'' in~\cite{KHeCVPR2020}). The slow-progressing property of the additional encoder is acquired by updating via a moving average from the main encoder with a large momentum.

To enable the usage of a large number of negative samples for training a re-ID model, we devise Momentum ReID (MoReID) architecture that is constructed with a Siamese network and a dictionary from MoCo. During training, a batch consisting of multiple groups, where the samples in the same group share the same ID, is fed to both networks. Only the outputs of the secondary \emph{forward-only} encoder with their corresponding IDs are stored in the dictionary. We treat each encoded sample from the main network as the query, which is compared with each of the keys in the dictionary to validate whether they match or not. All the keys with IDs different from the query are considered as negative samples. Among the keys that share the same ID with the query, only the samples that just came in from the current batch are used as positive samples (i.e., present positives) for training. Past positives are not used. While processing a batch of 256 samples, MoReID can leverage up to 8192 negative samples accumulated within the dictionary when using 4 GPUs with 48G GPU memory.

To take full advantage of a very large number of negative samples accumulated within the MoReID architecture, we use a novel loss called \emph{Hard-distance Elastic Loss} (HE Loss). Previous re-ID methods have mainly used the Triplet loss~\cite{EHofferSIMBAD2015,FSchroffCVPR2015,AHermansArXiv2017} which computes the difference between the distance to a single positive representation and the distance to a single negative representation for a given query (left illustration of Figure~\ref{fig:HE_Loss}). A single representation is an average feature over all samples or hard samples in each group. As the number of samples increases, the representativeness of a single representation gets less accurate. Because of this, it does not scale well with the sample size.

Instead of treating a group of samples with a single representation, the HE loss has the capability of leveraging multiple hard samples that are on the other side of the boundary optimally drawn between the positive and negative samples (i.e., positives on the negative side and negatives on the positive side). In detail, the HE loss is calculated as the sum of the penalties, which is the extent to which the boundary is crossed for all hard samples (right illustration of Figure~\ref{fig:HE_Loss}). For each query, the boundary is optimally determined to minimize the loss. We have theoretically demonstrated that it is optimal to place the boundary in the region where the number of hard positive samples and the number of hard negative samples are equal.

To demonstrate the effectiveness of the MoReID architecture and the HE loss, we carried out experiments on person re-ID and vehicle re-ID tasks. Experiments show that deploying only one of the two modules does not bring any advantage in improving re-ID accuracy. However, using the two modules together provides drastic improvement in accuracy when compared to the baseline without the modules in all re-ID tasks. The two novel modules work mutually and in a complementary way, meaning that the large negative training samples acquired using the MoReID architecture cannot be used effectively without the HE loss, whereas HE loss can only demonstrate its full strength when the number of `negative samples are at large'.

\section{Related Works}
\label{sec:rel_works}

\noindent{\bf ReID loss.} Existing re-ID methods typically use two types of losses with different label forms: pairwise (i.e., class-level label and model estimate) loss and tripletwise (i.e., query, and positive and negative samples for the query) loss. A pairwise loss defined to minimize the difference between the label and estimate is mainly used for ID classification in re-ID methods. Tripletwise loss is used in re-ID methods to bring samples with the same ID closer together and push the samples with different IDs further away from each other.

For the pairwise loss, there are several commonly used losses, e.g., softmax~\cite{FWangSPL2017,HWangCVPR2018}, cross-entropy loss~\cite{PChenICCV2021,XHaoICCV2021,MLiICCV2021,HParkICCV2021}, cosface~\cite{HWangCVPR2018}, arcface loss~\cite{JDengCVPR2019}, and circle loss~\cite{YSunCVPR2020,JZhaoICCV2021}. On the other hand, Triplet loss~\cite{AHermansArXiv2017} is the most representative of the tripletwise loss in the re-ID methods. Triplet loss measures the relationship between a query sample and two single representative samples representing the positives and the negatives, respectively. Each representation can be calculated using either a weighted sum of all samples in each group~\cite{XHaoICCV2021,YHuangICCV2021} or hard sample mining~\cite{LZhangICCV2019,YSunCVPR2020,AAichICCV2021,MLiICCV2021,HParkICCV2021}. However, as the number of samples increases significantly, it becomes more difficult to adequately represent them with a single representation, which we call the sample scalability issue. Other metric learning losses that can be used as a tripletwise loss also face similar limitations, e.g., c-triplet loss~\cite{FWangACMMM2017}, circle loss~\cite{YSunCVPR2020,JZhaoICCV2021}, CDF-based weighted triplet loss~\cite{LZhangICCV2019}, etc. To address sample scalability, N-pair loss~\cite{KSohnNeurIPS2016} and Ranked List loss~\cite{XWangCVPR2019} are defined to use multiple hard samples as representative samples, but show lower re-ID performance than Triplet loss because the number of hard samples is not optimally determined. In this paper, we introduce a novel loss, referred to as \emph{Hard-distance Elastic Loss} (HE Loss), that can better cope with the sample scalability issue to exhibit significantly higher accuracy than the Triplet loss.\smallskip

\noindent{\bf Exploring a large number of negative samples.} Recently, exploration of a very large number of negative samples in training has shown promising accuracy in various unsupervised representation learning tasks. Most of the methods presenting such promising accuracy employ the MoCo framework~\cite{KHeCVPR2020}, which provides a structural way to generate  a large number of negative samples. This MoCo framework has also been used for several Re-ID methods, but is only limited to unsupervised learning settings~\cite{HChenICCV2021,DFuCVPR2021,KZhengCVPR2021,YZhengICCV2021}. For the task of supervised re-ID, our MoReID is the first architecture to adopt MoCo to leverage a large number of negative samples with the help of the proposed HE loss.

\section{Method}
\label{sec:method}

\subsection{Hard-distance Elastic Loss}
\label{ssec:he_loss}

\noindent{\bf Definition of HE loss.} Hard-distance Elastic (HE) loss measures the degree to which samples cross the boundary (i.e., positive samples located on the negative side, and vice versa) that is determined to divide the feature space into positive and negative regions. Here, samples in the opposite side of the boundary are considered hard samples.

For a query $q$, key samples are split into its positives $p\in P$ and its negatives $n\in N$ according to their IDs, where $P$ and $N$ are the positive and negative sets, respectively. 
HE loss, $\mathcal{L}_{q}$, is defined with the boundary $t$ as follows:
\begin{equation}
  \mathcal{L}_{q}(t)=\sum_{p\in P}{\max(d_{pq}-t,0)}+\sum_{n\in N}{\max(t-d_{nq},0)},\label{eq:he_loss}
\end{equation}
where $d_{pq}$ and $d_{nq}$ are the distances (e.g., Euclidean distance in our experiments) from $q$ to $p$ and from $q$ to $n$, respectively. The optimal boundary $t^*$ can be acquired by minimizing the HE loss as follows:
\begin{equation}
    t^* = \argmin_t{\mathcal{L}_q{(t)}}.
\end{equation}

\noindent{\bf Derivation of the optimal boundary.} The optimal point of the boundary to minimize the HE loss must satisfy two conditions: i) the derivative of the HE loss with respect to the boundary $t$ is zero at this optimal point, and ii) the HE loss is convex with respect to the boundary $t$. The derivative of $\mathcal{L}_q(t)$ with respect to the boundary $t$ can be derived from eq.~\ref{eq:he_loss} as follows:
\begin{equation}
    \frac{d\mathcal{L}_q(t)}{dt}=\sum_{p\in P}{-\mathds{1}(d_{pq}> t)}+\sum_{n\in N}{\mathds{1}(d_{nq}< t)}=-N_{hp}(t)+N_{hn}(t)\label{eq:derivative_he_loss}
\end{equation}
where $\mathds{1}(\cdot)$ is a unit function. $N_{hp}(t)$ and $N_{hn}(t)$ are the number of hard positive samples and hard negative samples, respectively, when the boundary is $t$. 
Note that $\mathds{1}(d_{pq}> t)$ or $\mathds{1}(d_{nq}< t)$ represent samples that are located on the other side of the boundary with respect to its identity. In other words, the samples satisfying the constraints within the unit functions in eq.~\ref{eq:derivative_he_loss} are the hard samples. Accordingly, the first condition to define an optimal boundary is satisfied when the number of hard positive samples and the number of hard negative samples are equal (i.e., $N_{hp}$ = $N_{hn}$).

In addition, $N_{hp}(t)$ is a monotonically decreasing function for $t$ as the number of hard positive samples decreases as the boundary moves away from the query. Similarly, $N_{hn}(t)$ is a monotonically increasing function. Accordingly, $d\mathcal{L}_q(t)/dt$ in eq.~\ref{eq:derivative_he_loss} is a monotonically increasing function and the HE loss is convex with respect to $t$, satisfying the second condition for an optimal boundary.

Knowing that the second condition is always met, the task of finding the optimal boundary is narrowed down to localizing the point where $N_{hp}$ = $N_{hn}$.
We first compute the distances from all the samples to the query and sort them in ascending order. Then the shortest distance is set to be the first boundary candidate to evaluate whether $N_{hp}$ = $N_{hn}$ is satisfied. If satisfied, the candidate is set as $t^*$. If not, we iterate through the sorted distances until the condition is met. HE loss for a given batch is acquired by processing all the queries and their optimal boundaries independently and averaging the results. Pseudo-code for computing the optimal boundary and the HE loss is included in Algorithm~\ref{alg:moreid_pseudo}.

\subsection{MoReID}
\label{ssec:architecture}

\noindent{\bf Architecture.} A core component of MoReID is being able to use a very large number of negative samples in training a re-ID model. To enable this capability, MoReID leverages the MoCo architecture~\cite{KHeCVPR2020}, which is equipped with a dictionary that continuously collects a large number of encoded features for incoming input images as the training evolves. The dictionary is a queue where the oldest batch leave as a new batch comes in.

\begin{wrapfigure}{R}{0.45\textwidth}
    \captionsetup{font=small}
    \centering
    \includegraphics[trim=5mm 5mm 5mm 5mm,clip,width=\linewidth]{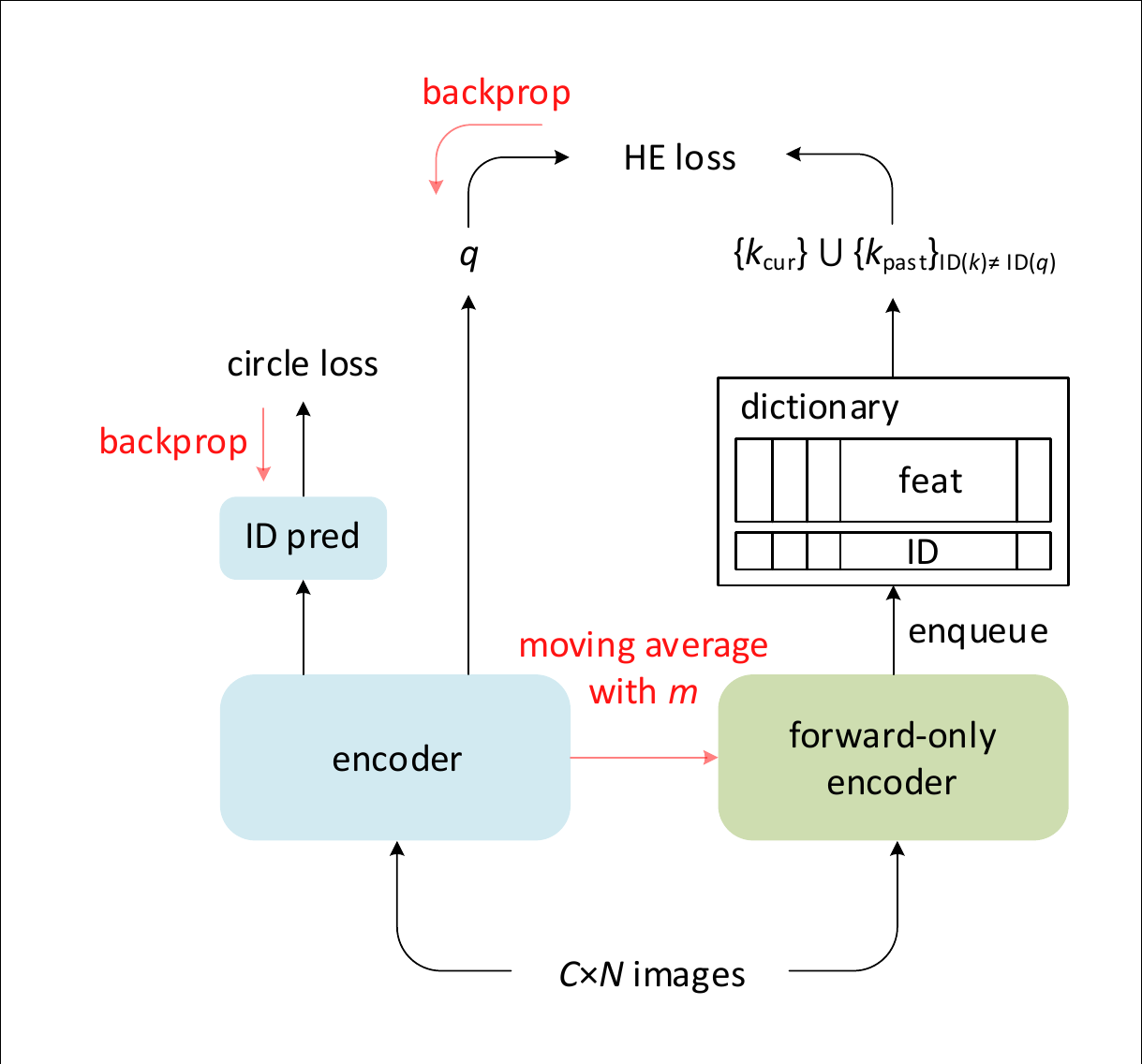}
    \vspace{-0.6cm}
    \caption{{\bf MoReID.} Black arrows and red arrows represent forward computations and model updates, respectively.}
    \label{fig:architecture}
\end{wrapfigure}

To take full advantage of the benefits of using a large number of samples (especially, when generating the precise boundary between positive and negative samples in the feature space), it is desirable that these samples be encoded into a consistent feature space. To this end, MoCo uses a separate, slow-progressing encoder in addition to the main encoder to encode the samples to lie in a nearly consistent feature space. Note that this slow-progressing encoder only encodes the key samples that are fed into the dictionary. 

While the main encoder is updated via back-propagation, this slow-progressing encoder (``forward-only encoder'' in figure~\ref{fig:architecture}) is updated via a moving average with momentum $m$ from the main encoder (i.e., $f_k=m\cdot f_k+(1-m)\cdot f_q$, where $f_k$ and $f_q$ are the slow-progressing encoder and the main encoder, respectively). The slow progress property in the forward-only encoder is acquired by reducing the extent of update with a large momentum $m$. Note that the forward-only encoder is only used for training, as its purpose is to provide a large number of samples (negative samples for our method) to be compared against a given query.

As a result, our MoReID architecture is also designed as a Siamese network that consists of the main encoder and the additional forward-only encoder. Both encoders take the same image batch as input where each image batch consists of $C$ groups, and each group contains $N$ images with the same ID. The main encoder is trained with two losses: pairwise ID loss and the proposed HE loss. For ID classification, the output of the main encoder is fed through an ID prediction layer which generates the input to the computation of the ID loss. We use circle loss~\cite{YSunCVPR2020} to serve as the ID loss\footnote{Circle loss~\cite{YSunCVPR2020} is designed in two types, pairwise and tripletwise, and we adopt the pairwise type as the ID loss in MoReID architecture.}. The proposed MoReID structure is shown in Figure~\ref{fig:architecture}.\smallskip

\noindent{\bf Labeling key samples in the dictionary.} Within the dictionary, only the samples that do not share the same ID with a given query are treated as negative samples. As for the positive samples (samples with same ID w.r.t the query), only the samples from the current batch is considered, while disregarding the ones that had been previously stored in the dictionary. This was an empirical decision. The experiments demonstrated that leaving out the past positive samples led to higher accuracy, and details on this are found in the supplementary material. If a model is trained with past positive samples, the model is less likely to carry the capability to properly separate the positives from the negatives in the current feature space. This may be because the past positive samples do not share the consistent feature space with the query. 
As no past positive samples are used for training, only \emph{negative samples are at large}.

As the samples within the dictionary are constantly picked out based on their IDs, the dictionary is constructed so as to contain both the features and their IDs. The \texttt{LabelDictionary} function in Algorithm~\ref{alg:moreid_pseudo} deals with the labeling process (positive or negative w.r.t. the query) described in this section.

\begin{algorithm}[t]
\SetAlCapFnt{\small}
\caption{{\small MoReID: PyTorch-like Pseudo-code}}\label{alg:moreid_pseudo}
{\scriptsize
\begin{tabular}{l}
\hspace{-0.5cm}\textcolor{myblue}{\# f\_q, f\_k: main and secondary encoders}\\
\hspace{-0.5cm}\textcolor{myblue}{\# h: ID pred. layer}\\
\hspace{-0.5cm}\textcolor{myblue}{\# feat, id: dictionary for feat and ID}\\
\hspace{-0.5cm}\textcolor{myblue}{\# m: momentum}\\
\hspace{-0.5cm}~~~~\\
\hspace{-0.5cm}f\_k.params = f\_q.params \textcolor{myblue}{\# initialize}\\
\hspace{-0.5cm}\textcolor{myblue}{\# load a minibatch (image and label)}\\
\hspace{-0.5cm}\textcolor{myred}{for} x, l \textcolor{myred}{in} loader:\\
\hspace{-0.5cm}~~~~Q, K = f\_q(x), f\_k(x) \textcolor{myblue}{\# encode}\\
\hspace{-0.5cm}~~~~P = h(Q) \textcolor{myblue}{\# predict ID}\\
\hspace{-0.5cm}~~~~K = K.detach() \textcolor{myblue}{\# no gradient to f\_k}\\
\hspace{-0.5cm}~~~~\\
\hspace{-0.5cm}~~~~dequeue(feat) \textcolor{myblue}{\# dequeue the earliest feat}\\
\hspace{-0.5cm}~~~~enqueue(feat, K) \textcolor{myblue}{\# enqueue the cur feat}\\
\hspace{-0.5cm}~~~~dequeue(id) \textcolor{myblue}{\# dequeue the earliest id}\\
\hspace{-0.5cm}~~~~enqueue(id, l) \textcolor{myblue}{\# enqueue the cur id}\\
\hspace{-0.5cm}~~~~\\
\hspace{-0.5cm}~~~~\textcolor{myblue}{\# loss (closs: circle loss)}\\
\hspace{-0.5cm}~~~~loss = heloss(Q, feat, l, id) + closs(P, l)\\
\hspace{-0.5cm}~~~~\\
\hspace{-0.5cm}~~~~\textcolor{myblue}{\# update MoReID}\\
\hspace{-0.5cm}~~~~loss.backward()\\
\hspace{-0.5cm}~~~~f\_k.params = m*f\_k.params\\
\hspace{-0.5cm}~~~~~~~~~~~~~~~~~~~~~~~~~~+(1-m)*f\_q.params\\
\\
\\
\\
\\
\\
\\
\\
\\
\\
\end{tabular}
\begin{tabular}{l}
\textcolor{myblue}{\# HE loss}\\
\textcolor{myred}{def} \textcolor{mygreen}{heloss}(Q, K, id\_Q, id\_K):\\
~~~~loss = 0\\
~~~~\textcolor{myred}{for} q, id\_q \textcolor{myred}{in} \textcolor{mygreen}{zip}(Q, id\_Q):\\
~~~~~~~~\textcolor{myblue}{\# label key samples in the dictionary}\\
~~~~~~~~\textcolor{myblue}{\# P, N: positive and negative keys}\\
~~~~~~~~P, N = LabelDictionary(K, id\_K, id\_q)\\
~~~~~~~~n\_p = P.size(0) \textcolor{myblue}{\# num of positives}\\
~~~~~~~~\textcolor{myblue}{\# calculate euclidean distance}\\
~~~~~~~~d\_pq, d\_nq = eucl(q, P), eucl(q, N)\\
~~~~~~~~\textcolor{myblue}{\# sort in ascending order}\\
~~~~~~~~d\_pq, d\_nq = d\_pq.sort(), d\_nq.sort()\\
~~~~~~~~\\
~~~~~~~~\textcolor{myblue}{\# search the optimal boundary}\\
~~~~~~~~\textcolor{myblue}{\# n\_hp, n\_hn: num of hard samples}\\
~~~~~~~~n\_hp, n\_hn = n\_p, 0\\
~~~~~~~~d\_pq\_cur, d\_nq\_cur = d\_pq[0], d\_nq[0]\\
~~~~~~~~\textcolor{myred}{while} 1:\\
~~~~~~~~~~~~\textcolor{myred}{if} d\_pq\_cur $<$= d\_nq\_cur:\\
~~~~~~~~~~~~~~~~t = d\_pq\_cur\\
~~~~~~~~~~~~~~~~n\_hp $-$= 1\\
~~~~~~~~~~~~~~~~d\_pq\_cur = d\_pq[n\_p-n\_hp]\\
~~~~~~~~~~~~\textcolor{myred}{else}:\\
~~~~~~~~~~~~~~~~t = d\_nq\_cur\\
~~~~~~~~~~~~~~~~n\_hn $+$= 1\\
~~~~~~~~~~~~~~~~d\_nq\_cur = d\_pq[n\_hn]\\
~~~~~~~~~~~~\\
~~~~~~~~~~~~\textcolor{myred}{if} n\_hp == n\_hn: \textcolor{myred}{break} \textcolor{myblue}{\# terminate}\\
~~~~~~~~\\
~~~~~~~~\textcolor{myblue}{\# calculate loss}\\
~~~~~~~~loss += relu(d\_pq - t).sum()\\
~~~~~~~~~~~~~~~~~~~~~~~~+ relu(t - d\_nq).sum()\\
~~~~\textcolor{myred}{return} loss/Q.size(0)
\end{tabular}
}
\end{algorithm}

\subsection{Implementation Details}
\label{ssec:implementation_details}

\noindent{\bf Encoder backbone.} For the encoder backbone design, we follow the optimal configuration of a Re-ID network in~\cite{LHeArXiv2020}. ResNet-50~\cite{KHeCVPR2015} is used which is followed by two non-local modules~\cite{XWangCVPR2018}. The output from the non-local modules are pooled by the generalized mean pooling (i.e., $\left(1/|\textbf{X}|\sum_{x\in \textbf{X}}{x^\alpha}\right)^{1/\alpha}$, where $\alpha$ = 3 in our experiments). All first batch normalizations in each residual module have been replaced by instance batch normalizations~\cite{DUlyanovCVPR2017}. The backbone network was initialized on ImageNet-pretrained ResNet-50. For ID prediction, one fully connected layer is used.\smallskip

\noindent{\bf Inference.} The re-ID task is to retrieve matching samples from the gallery for a given query. For the inference, the only main encoder is used to encode both query and gallery samples in the learned feature space. The similarity between a pair of encoded samples (query and gallery) is measured with the Euclidean distance.\smallskip

\noindent{\bf Optimization.} We use SGD as our optimizer. Weight decay and momentum are 0.0005 and 0.9, respectively. Base learning rate was set based on the dataset. We have conducted a study in the experiment section where we investigate the relationship between the dataset and the base learning rates. (see Sec.~\ref{ssec:study_of_lr}). The training schedule is commonly designed independent of the dataset so that the learning rate gradually decays to zero at the cosine rule beginning at the middle of the total training epoch. This is to gradually reduce the difference between the parameters of the main encoder and the forward-only encoder while training. Once the training is done, the key samples in the dictionary can be compared with their query sample in the `close-to-equivalent' feature space.

\section{Experiments}
\label{sec:experiments}

To confirm the effectiveness of the proposed re-ID method, we use three re-ID datasets: VeRi-776, Market-1501, and VeRi-Wild. Unless specified, all experiments (e.g., ablation studies, comparison of HE loss to other losses, etc.) were carried out on VeRi-776 dataset. Comparing our method with the state-of-the-art methods is performed on all three re-ID datasets. mAP and R-1 are commonly used as evaluation metrics for all the datasets.\smallskip

\noindent{\bf Training schedule.} For VeRi-776 and Market-1501, we trained for 60 epochs. For the large-scale VeRi-Wild dataset, we trained for 120 epochs.\smallskip

\noindent{\bf Pre-processing.} For training, three pre-processing approaches were found to be optimal after evaluating all possible combinations of multiple approaches in \cite{LHeArXiv2020}: horizontal flipping, random erasing~\cite{ZZongAAAI2020}, and auto-augment~\cite{ECubukCVPR2019}. For the vehicle re-ID datasets (i.e., VeRi-776 and VeRi-Wild), the input images are resized to 256$\times$256. For the person re-ID dataset (i.e., Market-1501), the input image is resized to 384 $\times$128.

\begin{table}[t]
\centering
\captionsetup{font=small}
\begin{minipage}{0.35\linewidth}
\includegraphics[trim=40mm 87mm 50mm 95mm,clip,width=\linewidth]{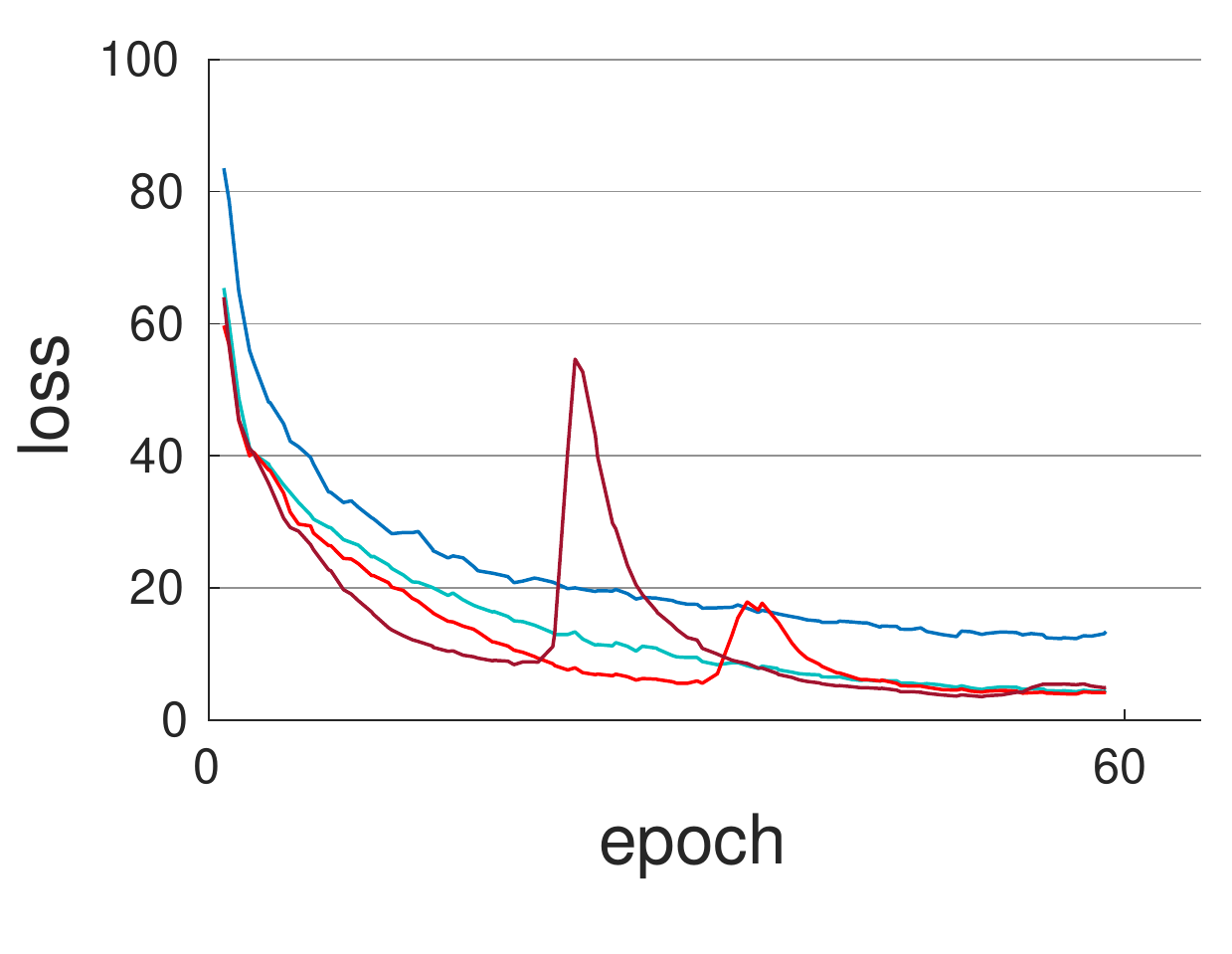}
\end{minipage}
\begin{minipage}{0.35\linewidth}
\includegraphics[trim=40mm 87mm 50mm 95mm,clip,width=\linewidth]{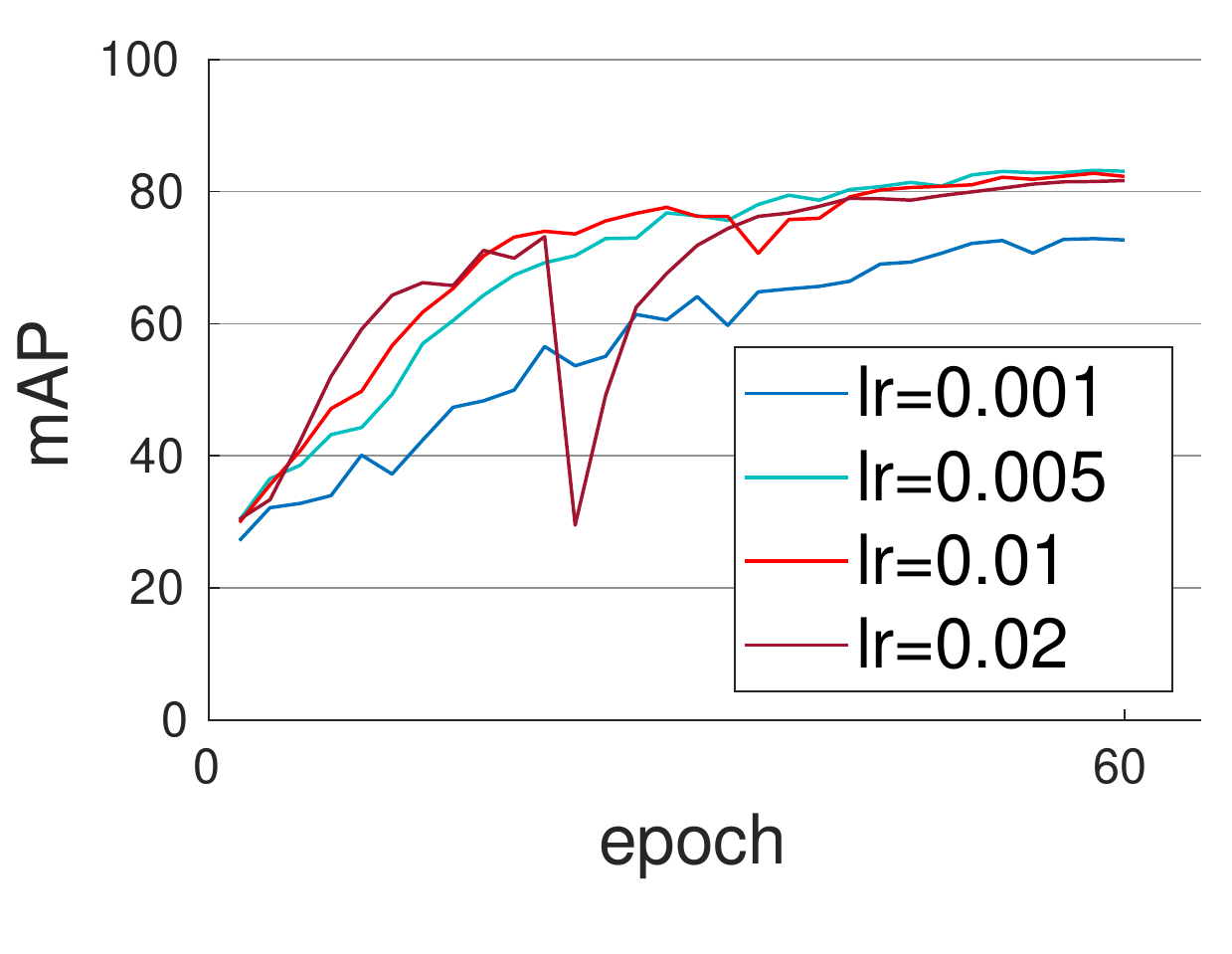}
\end{minipage}
~
~
\begin{minipage}{0.24\linewidth}
\setlength{\tabcolsep}{9.0pt}
\renewcommand{\arraystretch}{0.9}
{\small
\begin{tabular}{l|c}
\multicolumn{1}{c|}{lr} & mAP\\\specialrule{.15em}{.05em}{.05em}
0.001 & 72.7 \\
0.005 & 83.1 \\
0.01 & \bf{83.3} \\
0.02 & 81.7 \\
\multicolumn{2}{c}{}\\
\multicolumn{2}{c}{}\\
\multicolumn{2}{c}{}\\
\end{tabular}
}
\end{minipage}
\vspace{-0.2cm}
\caption{{\bf Comparisons of different learning rates.} The figures on the left and middle show the loss evolution and accuracy trend as training progresses, respectively. The table on the right shows mAPs of the models trained with different learning rates.}
\label{tab:learning_rate}
\end{table}

\subsection{Study for Optimal Learning Rate}
\label{ssec:study_of_lr}

\noindent{\bf Searching the optimal learning rate.} To search for the optimal learning rate for MoReID, we compare models trained with different learning rates as shown in Table~\ref{tab:learning_rate}. The optimal learning rate with respect to accuracy was 0.01. When the learning rate is lower (i.e., 0.001) than this optimal rate, the model was under-fitted, converging at a relatively higher loss with a degraded accuracy. 

On the other hand, when the learning rate was higher than or equal to the optimal rate, the training seemed unstable as a spike occurs in the middle of the training (Left figure in Table~\ref{tab:learning_rate}). The spike triggered a sudden drop in accuracy (Middle figure in Table~\ref{tab:learning_rate}). This spike in loss was also observed in~\cite{XChenICCV2021} when there was a sudden change in gradient at the initial layer of the model. Based on this observation, \cite{XChenICCV2021} provided a temporary solution for this instability by freezing the initial layer, which led to an increased accuracy. However, suppressing the gradient of the initial layer, including this layer freezing, gradient clipping, etc., was found to be ineffective in dealing with the instability within our scenario. More studies are needed to address this unique training instability in future works.\smallskip

\begin{table}[t]
\captionsetup{font=small}
\centering
\begin{minipage}{0.5\linewidth}
\includegraphics[trim=15mm 90mm 25mm 100mm,clip,width=\linewidth]{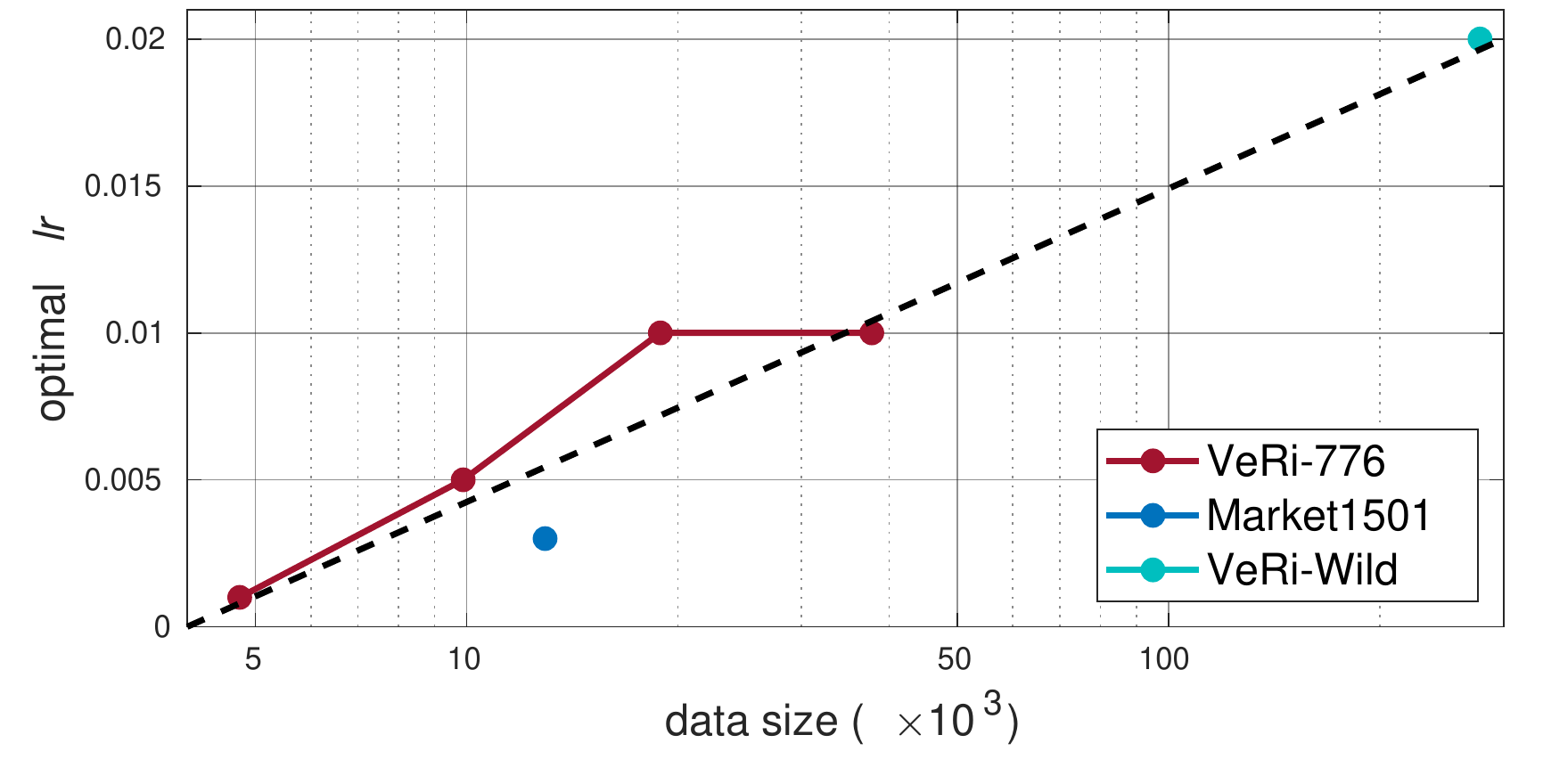}
\end{minipage}
\begin{minipage}{0.45\linewidth}
\setlength{\tabcolsep}{6.0pt}
\renewcommand{\arraystretch}{0.9}
{\small
\begin{tabular}{c|ccc}
& data size & $lr_s$ & mAP \\\specialrule{.15em}{.05em}{.05em}
\multirow{3}{*}{subsets} & 4,751 & 0.001 & 40.9 \\
& 9,889 & 0.005 & 56.7 \\
& 18,882 & 0.01 & 70.3 \\\hline
entire & 37,775 & 0.01 & 83.3\\
\multicolumn{4}{c}{}\\
\multicolumn{4}{c}{}\\
\end{tabular}
}
\end{minipage}
\caption{{\bf The optimal learning rate with respect to the size of the dataset.} The data size is expressed on a logarithmic scale along the x coordinate in the figure on the left. The dashed line in the figure is drawn to represent the approximate relationship between the optimal learning rate and the data size, expressed in eq.~\ref{eq:lr}. The data size, the optimal learning rate and accuracy for VeRi-776 and its three subsets are shown in the table on the right.}
\label{fig:optimal_lr}
\end{table}

\noindent{\bf The optimal learning rate with respect to the dataset.} As the accuracy of our method was somewhat sensitive to the learning rate, we saw the need to provide a study that shows the relationship between the optimal learning rate and a given dataset (especially the size of the dataset). This would prevent having to go through an exhaustive search just to find out the optimal learning rate when applying our method to a different dataset.

To figure out this relationship within the same dataset, we generate three subsets from VeRi-776 dataset by collecting samples each associated with 1/2, 1/4, and 1/8 of the entire IDs. Then, we check whether the relationship between the optimal learning rates of the entire VeRi-776 dataset and the three subsets were consistently applicable when we trained on other datasets (i.e., Market-1501 and VeRi-Wild). As shown in the figure of Table~\ref{fig:optimal_lr}, the optimal learning rate should increase logarithmically as the size of the dataset size gets larger and the relationship can simply be expressed as follows:
\begin{eqnarray}
    lr_s&=&0.02\frac{\log{(s)}-\log{(4\times 10^3)}}{\log{(3\times 10^5)}-\log{(4\times 10^3)}}\nonumber\\
    &\propto&\text{log}(s) ~~~\text{if}~~s \gg 4\times10^\text{3},\label{eq:lr}
\end{eqnarray}
where $s$ is the size of the dataset. Note that $\log{(3\times 10^5)}$ and $0.02$ in eq.~\ref{eq:lr} are the size of the largest dataset (i.e., VeRi-Wild) used to derive this equation and the optimal learning rate for this dataset, respectively. $lr_s$ is a reference value that can be used as the optimal learning rate when applying our method to a different dataset. From these observations, we can set a rule for searching the optimal $lr$ with respect to the dataset size as: \smallskip

\emph{``When the dataset size is increased by power of k, multiply the learning rate by k''}.\smallskip

\noindent As this rule does not conflict with the linear scale rule~\cite{PGoyalArXiv2018} that defines the optimal $lr$ with respect to the batch size, these two rules can be applied concurrently.

\subsection{Ablation Studies}
\label{ssec:ablation}

\noindent{\bf Momentum $m$.} We compare the accuracy of models trained with different momentum $m$ as shown in the table below:\smallskip

{\centering
\noindent
\setlength{\tabcolsep}{8.0pt}
\renewcommand{\arraystretch}{0.9}
{\small
\begin{tabular}{c|c|ccccc}
\centering
& \multirow{2}{*}{\textcolor{gray}{no dict.}} & \multicolumn{5}{c}{$m$} \\
 & & 0.9 & 0.99 & 0.995 & 0.997 & 0.999 \\\specialrule{.15em}{.05em}{.05em}
mAP & \textcolor{gray}{80.6} & 79.5 & 80.7 & 82.4 & \bf{83.3} & 83.0\\
gain & \textcolor{gray}{$\cdot$} & (-1.1) & \textcolor{teal}{(+0.1)} & \textcolor{teal}{(+1.8)} & \textcolor{teal}{(+2.7)} & ~\textcolor{teal}{(+2.4)}
\label{tab:dictionary_size}
\end{tabular}
}
\par}\smallskip

\noindent When $m$ is $\geq$ 0.995, the accuracy was reasonably high. This indicates that maintaining the key samples consistent by slowly updating the forward-only encoder was critical to achieving high accuracy. It is noteworthy that when consistency of samples was less maintained (i.e., $m$ $\leq$ 0.9), there was even a loss in accuracy when compared to ``no dict.''. \smallskip

\noindent{\bf Dictionary size.} We experiment how different dictionary sizes affect the mAPs. This comparison is shown in the table below:\smallskip

{\centering
\noindent
\setlength{\tabcolsep}{9.5pt}
\renewcommand{\arraystretch}{0.9}
{\small
\begin{tabular}{c|c|cccc}
\centering
& \multirow{2}{*}{\textcolor{gray}{no dict.}} & \multicolumn{4}{c}{dict. size} \\
 & & 1024 & 2048 & 4096 & 8192 \\\specialrule{.15em}{.05em}{.05em}
\# batch & \textcolor{gray}{$\cdot$} & 4 & 8 & 16 & 32 \\
optimal $m$ & \textcolor{gray}{$\cdot$} & 0.99 & 0.995 & 0.997 & 0.997 \\\hline
mAP & \textcolor{gray}{80.6} & 81.9 & 82.5 & 83.0 & \bf{83.3} \\
gain & \textcolor{gray}{$\cdot$} & \textcolor{teal}{(+1.3)} & \textcolor{teal}{(+1.9)} & \textcolor{teal}{(+2.4)} & ~\textcolor{teal}{(+2.7)}
\label{tab:dictionary_size}
\end{tabular}
}
\par}\smallskip

\noindent We have used a minibatch consisting of 256 images ($C$: 16, $N$: 16). ``$\#$ batch'' is the number of batches constantly stacked in a dictionary. Within our computing environment (4$\times$12GB GPUs), the maximum dictionary size is 8192. ``gain'' shows how much accuracy is gained by using a dictionary when compared with a model without a dictionary (``no dict.''). Regardless of the dictionary size, using a dictionary itself always provided better accuracy. In addition, increasing the dictionary size consistently bumped up the accuracy. 

How long a batch is kept in the dictionary affects the optimal $m$ that controls the speed of update for the forward-only encoder. More specifically, a slower update for the forward-only encoder (i.e., bigger $m$) may be needed to increase how long the past batches reside in the dictionary. We found the optimal $m$ for each dictionary size as shown in the table above. We have experimentally validated that as the dictionary size gets larger, the optimal $m$ also becomes larger.\smallskip

\noindent{\bf Batch.} HE loss is highly related to the number of hard samples. Since a large number of negative samples are used, the maximum number of hard samples actually becomes the twice the number of positive samples at the optimal boundary (i,e, $N_{hp}=N_{hn}$ in eq.~\ref{eq:derivative_he_loss}). As each batch is fixed with the configuration of $C$$\times$$N$ ($C$: number of ID types, $N$: number of instances per ID type), the number of positive samples is also fixed as $N$-1. Therefore, how the batch configuration is set directly affects the HE loss with respect to the accuracy as shown below:\smallskip

{\centering
\noindent
\setlength{\tabcolsep}{12.8pt}
\renewcommand{\arraystretch}{0.9}
{\small
\begin{tabular}{c|cccccc}
\centering
 $C$$\times$$N$ & 128$\times$2 & 64$\times$4 & 32$\times$8 & 16$\times$16 & 8$\times$32 \\\specialrule{.15em}{.05em}{.05em}
mAP & 77.9 & 79.7 & 83.0 & \bf{83.3} & 83.2
\label{tab:batch}
\end{tabular}
}
\par}\smallskip

\noindent As shown in the table, the accuracy improves when $N$ is larger than or equal to 8. We use the optimal configuration (16$\times$16) throughout all experiments.

\subsection{Main Results}
\label{ssec:main_results}

\noindent{\bf Synergy of HE loss and MoReID architecture.} Here, we demonstrate how each of the two new components impact accuracy and whether they are effectively complementary to each other. In Table~\ref{tab:new_component}, we compare the accuracy of four different combinations with and without each component on top of the baseline. The baseline uses the same backbone network as ours, while Triplet loss are used instead of the HE loss. While using HE loss (method (a)) slightly improves the baseline accuracy by 0.9, introducing a large number of negative samples using MoReID architecture (method (b)) rather brings a degraded accuracy. When the model is equipped with both of the components, a synergistic effect can be observed where the accuracy is significantly increased over the baseline.

The result can be interpreted in three aspects: i) Being able to represent positive or negative groups with multiple hard samples (i.e., using the HE loss) has a benefit over using a single representation for each group, ii) Without a proper loss design, using a large number of samples has very little impact on accuracy improvement, and that iii) the two components effectively complement each other.\smallskip

\begin{table}[t]
\setlength{\tabcolsep}{17.8pt}
\renewcommand{\arraystretch}{0.9}
\centering
{\small
\begin{tabular}{c|cc|c}
method & MoReID & HE loss & mAP \\\specialrule{.15em}{.05em}{.05em}
\textcolor{gray}{baseline} & & & \multicolumn{1}{l}{\textcolor{gray}{79.7}} \\
(a) & & \cmark & \multicolumn{1}{l}{80.6 \textcolor{teal}{(+0.9)}} \\
(b) & \cmark & & \multicolumn{1}{l}{79.4 (-0.3)} \\
{\bf Ours} & \cmark & \cmark & \multicolumn{1}{l}{{\bf 83.3} \textcolor{teal}{(+3.6)}} \\
\end{tabular}
}
\caption{{\bf Accuracy with and without MoReID and HE loss, respectively.} All methods use the same training schedule. `baseline' and `method (a)' use a learning rate of 0.01 which has been proven to be optimal for the backbone network as in~\cite{LHeArXiv2020}.}
\label{tab:new_component}
\end{table}

\noindent{\bf Comparison to the state-of-the-arts.} Table~\ref{tab:comparison_with_previous_methods} shows the results on three re-ID datasets. Our method consistently outperforms the baseline and all previous methods on all three datasets in terms of both mAP and R-1. 

On {\bf VeRi-776}, our method increases baseline accuracy by 3.3 in mAP and 0.5 in R-1. Our method also presents a better mAP by 0.2 while yielding a comparable R-1 compared to the SOTA (i.e., HRC~\cite{JZhaoICCV2021}).

On {\bf Market-1501}, we increases baseline accuracy by 2.6 in mAP and 0.8 in R-1. Our method outperforms the SOTA in mAP by 0.5 and presents a comparable R-1 performance.

On {\bf VeRi-Wild}, our method increases baseline accuracy in all three subsets. The improvements are $\sim$2.6 and $\sim$0.6 in mAP and R-1, respectively. Our method outperforms the SOTA by $\sim$1.9 in mAP and by $\sim$3.7 in R-1, in all three subsets.

Our method consistently provides state-of-the-art accuracy for both vehicle re-ID and person re-ID tasks. This result also demonstrates that our method is scalable to a large-scale dataset (i.e., VeRi-Wild dataset). It is noteworthy to mention that the gain over the SOTA was higher on the large-scale dataset than that on other datasets.

\begin{table*}[t]
\captionsetup{font=small}
\subfloat[{\small VeRi-776}]{
\setlength{\tabcolsep}{1.0pt}
\renewcommand{\arraystretch}{0.9}
\centering
{\scriptsize
\begin{tabular}[t]{l|cc}
&&\\
method & mAP & R-1 \\
\specialrule{.15em}{.05em}{.05em}
CAL~\cite{YRaoICCV2021} & 74.3 & 95.4 \\
PGAN~\cite{XZhangTITS2020} & 79.3 & 96.5 \\
PVEN~\cite{DMengCVPR2020} & 79.5 & 95.6 \\
SAVER~\cite{PKhorramshahiECCV2020
} & 79.6 & 96.4 \\
HRC~\cite{JZhaoICCV2021} & 83.1 & {\bf 97.3} \\\hline
baseline & 80.0 & 96.8 \\
Ours & {\bf 83.3} & {\bf 97.3} \\
gain & \textcolor{teal}{+3.3} & \textcolor{teal}{+0.5}\\
\end{tabular}
\label{tab:veri776}
}
}
\hspace{-0.3cm}
\subfloat[{\small Market-1501}]{
\setlength{\tabcolsep}{1.0pt}
\renewcommand{\arraystretch}{0.9}
\centering
{\scriptsize
\begin{tabular}[t]{l|cc}
&&\\
method & mAP & R-1 \\
\specialrule{.15em}{.05em}{.05em}
PAT~\cite{YLiCVPR2021} & 88.0 & 95.4 \\
ABD~\cite{TChenICCV2019} & 88.3 & 95.6 \\
BV~\cite{CYanICCV2021v2} & 89.1 & 96.0 \\
SSGR~\cite{CYanICCV2021} & 89.3 & {\bf 96.1} \\
CAL~\cite{YRaoICCV2021} & 89.5 & 95.5 \\\hline
baseline & 87.4 & 95.3 \\
Ours & {\bf 90.0} & {\bf 96.1} \\
gain & \textcolor{teal}{+2.6} & \textcolor{teal}{+0.8}\\
\end{tabular}
\label{tab:Market1501}
}
}
\hspace{-0.3cm}
\subfloat[{\small VeRi-Wild}]{
\setlength{\tabcolsep}{1.0pt}
\renewcommand{\arraystretch}{0.9}
\centering
{\scriptsize
\begin{tabular}[t]{l|cc|cc|cc}
\multirow{2}{*}{method} & \multicolumn{2}{c|}{small} & \multicolumn{2}{c|}{medium} & \multicolumn{2}{c}{large} \\
& mAP & R-1 & mAP & R-1 & mAP & R-1 \\
\specialrule{.15em}{.05em}{.05em}
BW~\cite{RKumarIJCNN2019} & 70.5 & 84.2 & 62.8  & 78.2 & 51.6 & 70.0 \\
SAVER~\cite{PKhorramshahiECCV2020} & 80.9 & 94.5 & 75.3 & 92.7 & 67.7 & 89.5 \\
PVEN~\cite{DMengCVPR2020} & 82.5 & $\cdot$ & 77.0  & $\cdot$ & 69.7 & $\cdot$ \\
PGAN~\cite{XZhangTITS2020} & 83.6 & 95.1 & 78.3 & 92.8 & 70.6 & 89.2 \\
HRC~\cite{JZhaoICCV2021} & 85.2 & 94.0 & 80.0 & 91.6 & 72.2 & 88.0 \\\hline
baseline & 83.7 & 95.5 & 78.8 & 94.4 & 71.5 & 91.5 \\
Ours & {\bf 86.1} & {\bf 96.1} & {\bf 81.2} & {\bf 94.5} & {\bf 74.1} & {\bf 91.7} \\
gain & \textcolor{teal}{+2.4} & \textcolor{teal}{+0.6} & \textcolor{teal}{+2.4} & \textcolor{teal}{+0.1} & \textcolor{teal}{+2.6} & \textcolor{teal}{+0.2}\\
\end{tabular}
\label{tab:veri_wild}
}
}
\caption{{\bf Comparison with previous methods (top 5) on three datasets.} Numbers in the parentheses indicate the performance gains with respect to the corresponding baselines. For fair comparison, all listed methods do not use any post-processing such as re-rank nor any external datasets.}
\label{tab:comparison_with_previous_methods}
\end{table*}

\subsection{Tripletwise ReID Loss Comparison}
\label{ssec:loss_comparison}

We compare the proposed HE loss with other tripletwise losses used to define triplet relationships beyond accuracy.\smallskip

\noindent{\bf vs. Triplet loss.} Figure~\ref{fig:HE_Loss} shows the conceptual difference between the HE loss and the Triplet loss. The most distinct property of HE loss compared to Triplet loss is that it can take more than one samples to represent the positive and the negative sets. As previously claimed, a single representation used by Triplet loss cannot cope well with the scalability of samples, which must be considered when using MoReID. To confirm this claim, we compare the accuracy of HE loss and Triplet loss while changing the dictionary size as shown in the Table below:\smallskip

{\centering
\noindent
\setlength{\tabcolsep}{13.2pt}
\renewcommand{\arraystretch}{0.9}
{\small
\begin{tabular}{c|cccccc}
\centering
\multirow{2}{*}{loss} & \multicolumn{5}{c}{dict. size} \\
 & 512 & 1024 & 2048 & 4096 & 8192 \\\specialrule{.15em}{.05em}{.05em}
Tri-all & 79.7 & 79.6 & 79.5 & 79.5 & 79.3 \\
Tri-hard & 80.2 & 79.9 & 79.7 & 79.5 & 79.7 \\
{\bf HE loss} & {\bf 81.0} & {\bf 81.9} & {\bf 82.5} & {\bf 83.0} & ~{\bf 83.3}
\label{tab:vs_triplet_loss}
\end{tabular}
}
\par}\smallskip

\noindent Margin of Triplet loss is set to 0.3 as in~\cite{LHeArXiv2020}. When a single representation is computed for either the positive or the negative sample set, two different Triplet losses can be deployed based on the fact whether hard sample mining precedes (``Tri-hard'') or not (``Tri-all''). We can observe the benefit of using the HE loss when using a larger-scale samples, whereas using the Triplet loss does not provide such advantage. Seeing that the ``Tri-hard'' slightly outperforms the ``Tri-all'' shows that the hard sample mining, which HE loss is also inherently leveraging, is more apt in handling the scalability of samples.

There exist several Triplet loss variants which also rely on single representations for the positive and negative sets, e.g., circle triplet loss (Circle)~\cite{YSunCVPR2020}, classification version of triplet loss (C-triplet)~\cite{FWangACMMM2017}, and CDF-based weighted triplet loss (W-triplet)~\cite{LZhangICCV2019}. We compare the accuracy of the proposed HE loss with these variants when used for MoReID training, as below:\smallskip

{\centering
\noindent
\setlength{\tabcolsep}{22.0pt}
\renewcommand{\arraystretch}{0.9}
{\footnotesize
\begin{tabular}{ccc|c}
\centering
Circle & C-triplet & W-triplet & {\bf HE} \\\specialrule{.15em}{.05em}{.05em}
79.1 & 80.3 & 79.9 & {\bf 83.3}
\label{tab:vs_triplet}
\end{tabular}
}
\par}\smallskip

\noindent Dictionary size of 8192 is used consistently for all losses. HE loss presents significantly better accuracy than the other losses, demonstrating the HE loss's ability to effectively process large-scale negative samples in training.\smallskip

\noindent{\bf vs. Losses that allows for multiple negatives.} Some previous losses have provided a way to represent an entire set of samples with multiple samples to better cope with large-scale negative samples. For example, N-pair loss~\cite{KSohnNeurIPS2016} can use multiple hard samples to represent all negative samples and Ranked List loss~\cite{XWangCVPR2019} can use multiple hard samples to represent both positive and negative samples. To compare HE loss with N-pair and Ranked List losses in handling the large-scale negative samples, we have set up an experiment to optimize two frameworks, the backbone network (our baseline in Table~\ref{tab:new_component}) and the MoReID network (dict. size is 8192). While the baseline processes all triplet combinations within the current batch only, the MoReID network also leverages past samples to be used in building a larger-scale negative sample set.\smallskip

{\centering
\noindent
\setlength{\tabcolsep}{3.5pt}
\renewcommand{\arraystretch}{0.9}
{\footnotesize
\begin{tabular}{c|cccc}
\centering
re-ID framework & \textcolor{gray}{Triplet} & N-pair & Ranked List & {\bf HE} \\\specialrule{.15em}{.05em}{.05em}
baseline & \multicolumn{1}{l}{\textcolor{gray}{79.7}} & \multicolumn{1}{l}{78.8} & \multicolumn{1}{l}{78.8} & \multicolumn{1}{l}{{\bf 80.6}} \\
MoReID & \multicolumn{1}{l}{\textcolor{gray}{79.4 (-0.3)}} & \multicolumn{1}{l}{79.8 \textcolor{teal}{(+1.0)}} & \multicolumn{1}{l}{80.5 \textcolor{teal}{(+1.7)}} & \multicolumn{1}{l}{{\bf 83.3 \textcolor{teal}{(+2.7)}}}
\label{tab:vs_multi_negative_loss}
\end{tabular}
}
\par}\smallskip

\noindent Numbers in parentheses indicate gaps from its corresponding baseline. N-pair loss and Ranked List loss present better accuracy with MoReID than with their baselines, demonstrating their ability to properly handle large-scale negative samples. Interestingly, both losses underperformed the Triplet loss when with the baselines, but outperformed it with MoReID. Remarkably, HE loss yields the best accuracy as well as the highest margin from its baseline. It shows the superiority of HE loss with regard to re-ID performance and its ability to process large-scale negative samples.\smallskip

\noindent{\bf vs. InfoNCE loss.} InfoNCE~\cite{AOordArXiv2018} is a widely used loss function in self-supervised learning that has been actively studied recently. InfoNCE also involves the triplet relationships which is formulated as shown below:
\begin{equation}
    \mathcal{L}_q^{i} = -\text{log}{\frac{\text{exp}(q\cdot p / \tau)}{\sum_{k\in \text{{\bf K}}} \text{exp}(q\cdot k / \tau)}},
    \label{eq:infonce}
\end{equation}
where $p$ is a positive sample and $k\in\text{{\bf K}}$ is a key sample. $\tau$ is a temperature parameter (set as 0.07 following ~\cite{ZWuCVPR2018}). Two distinct properties of InfoNCE when compared with the HE loss are: 1) only a single positive sample is considered, and 2) all key samples are used in the loss calculation.

Since this loss cannot be directly applied to our scenario where multiple positive examples exist, it should be modified to account for multi-label cases where multiple positive examples are allowed. \cite{PKhoslaNeurIPS2020} showed that multi-labeled InfoNCE can be designed in two ways\footnote{In fact, most methods that use InfoNCE which is modified for supervised learning use one of these two. \cite{AMiechCVPR2020} used $\mathcal{L}_q^{i,in}$ while \cite{BGunelICLR2021,MZhengICCV2021,HLeeICASSP2022} used $\mathcal{L}_q^{i,out}$.} to acquire this ability as:
\begin{eqnarray}
&\mathcal{L}_q^{i,in} =& -\text{log}{\sum_{p \in \text{{\bf P}}_q}{\left(\frac{\text{exp}(q\cdot p / \tau)}{\sum_{k\in \text{{\bf K}}} \text{exp}(q\cdot k / \tau)}\right)}},\label{eq:infonce_in}\\
&\mathcal{L}_q^{i,out} =& -\sum_{p \in \text{{\bf P}}_q}{\left(\text{log}{\frac{\text{exp}(q\cdot p / \tau)}{\sum_{k\in \text{{\bf K}}} \text{exp}(q\cdot k / \tau)}}\right)},\label{eq:infonce_out}
\end{eqnarray}
where $\text{{\bf P}}_q$ is the set of positive samples for the query $q$. 

On top of the vanilla InfoNCE, we have generated several variants by adjusting one or both of the ``two distinct properties'' as shown in the table below:\smallskip

{\centering
\noindent
\setlength{\tabcolsep}{15.0pt}
\renewcommand{\arraystretch}{0.9}
{\small
\begin{tabular}{c|c|c|c}
method & $C$$\times$$N$ & w/ hard mining & mAP \\\specialrule{.15em}{.05em}{.05em}
InfoNCE (eq.~\ref{eq:infonce}) & 128$\times$2 & no & 62.9 \\
InfoNCE (eq.~\ref{eq:infonce}) & 128$\times$2 & yes & 63.4 \\
HE & 128$\times$2 & & 77.5 \\\hline
InfoNCE (eq.~\ref{eq:infonce_in}) & 16$\times$16 & no & 70.5 \\
InfoNCE (eq.~\ref{eq:infonce_out}) & 16$\times$16 & no & 72.8 \\
InfoNCE (eq.~\ref{eq:infonce_in}) & 16$\times$16 & yes & 70.9 \\
InfoNCE (eq.~\ref{eq:infonce_out}) & 16$\times$16 & yes & 73.1 \\
HE & 16$\times$16 & & ~83.3
\label{tab:vs_infonce}
\end{tabular}
}
\par}\smallskip

\noindent First of all, instead of using only a single positive sample ($N$=2), we can adjust the model to take more than one positive samples ($N$=16). Another form of variation is to feed InfoNCE with a selected number of key samples instead of using all of them for loss calculation. This variation is labeled as ``w/ hard mining'' where the number of hard mined samples are set to be equivalent to the case of HE loss (i.e., 15). Note that the InfoNCE with no variation at all (i.e., vanilla InfoNCE) is listed as the first in the table. Results can be interpreted in three aspects: i) using hard negative samples selected by hard negative mining was effective in InfoNCE loss, ii) multi-label variant was also effective compared to the single-label one, and iii) all the variants of InfoNCE loss still underperformed HE loss by a significant margin.

\section{Discussion and Conclusion}
\label{sec:concl}

We have achieved the state-of-the-art accuracy on all three re-ID benchmarks by adopting the MoReID architecture and the HE loss. Wrapping up the experiments and analyses we have included in the paper, it is worthwhile to leave several discussion points. First, we came across an interesting phenomenon in training where the loss values appeared as spikes, especially when the learning rates are high. Further study that takes into account other training parameters (e.g., different optimizer, various batch sizes) might be helpful to unravel the phenomenon. Second, although the usage of InfoNCE loss has recently exploded in self-supervised learning, we have observed that it did not live up to its fame when it was used for supervised re-ID. It will be worthwhile to evaluate how HE loss will fit into the self-supervised learning paradigm.


\appendix

\section{Appendix}

\subsection{Details of Datasets Used for Evaluation}

In this section, we provide details of three re-ID datasets used for evaluation.

{\bf VeRi-776~\cite{XLiuECCV2016}} was developed for vehicle re-ID. This dataset consists of 49,357 images of 776 vehicles taken by 20 different cameras. 37,775 images with 576 IDs are used for training and the remaining images are used for testing. In the test set, 1,678 images are selected for the query set.

{\bf Market-1501~\cite{LZhengICCV2015}} was developed for person re-ID. It contains 32,668 images of 1,501 IDs. Images of each ID are taken by at most six cameras. 12,936 images and 19,732 images were used for training and testing, respectively. 3,368 images are selected as the query.

{\bf VeRi-Wild~\cite{YLouCVPR2019}} is a large-scale vehicle ID dataset which contains 416,314 images of 40,671 IDs from 174 cameras. 138,517 images of 10,000 IDs are used as a test set and evaluations are carried out with three subsets with 3,000 (\texttt{small}), 5,000 (\texttt{medium}), and 10,000 (\texttt{large}) IDs. The remaining images are used for training.

\subsection{Implementation: N-pair Loss and Ranked List Loss}

In the main manuscript, we chose the N-pair~\cite{KSohnNeurIPS2016} loss and the Ranked List loss~\cite{XWangCVPR2019} as the two representative losses that can leverage multiple negative samples. The original N-pair loss is developed for multi-class tasks and employs hard-negative class mining, which selects a pre-defined number of hard samples from each class except the class where the query belongs to. To be used as a tripletwise loss, however, we selected hard negative samples regardless of their classes, as negative representative samples. When evaluating the N-pair loss within the two re-ID frameworks (baseline and MoReID), we defined the number of hard negative samples to be 15 which was found to be optimal for both frameworks. As the N-pair loss does not allow multiple positives to be used as representative samples, only the hardest positive sample is leveraged.

Ranked List loss considers two sets of multiple hard examples to represent the positive and negative sets, respectively. Hard example mining is controlled by two parameters\footnote{Although the margin parameter is denoted as $m$ in \cite{XWangCVPR2019}, we use $\beta$ instead in this paper to avoid confusion with the momentum parameter.}, $\alpha$ and $\beta$ where the positives satisfying $d_{pq} > \alpha - \beta$ and the negatives satisfying $d_{nq} < \alpha$ are chosen as hard examples. When Ranked List loss is used with either baseline or MoReID, the optimal parameters were found to be 1.2 for $\alpha$ and 0.4 for $\beta$ after an exhaustive search.

\subsection{Ablation Study: Positive Samples Are Also at Large?} 

To validate our decision to only leverage the ``negative samples at large'' (i.e., excluding the past samples with the same ID as the query from training), we test out a ``positive samples are also at large" scenario where all excluded samples are pulled back in as positive samples. The comparison is shown below:\smallskip

{\centering
\noindent
\setlength{\tabcolsep}{16.5pt}
\renewcommand{\arraystretch}{0.9}
{\small
\begin{tabular}{c|ccc}
\centering
\multirow{2}{*}{w/ past positive samples} & \multicolumn{3}{c}{$m$} \\
 & 0.995 & 0.997 & 0.999 \\\specialrule{.15em}{.05em}{.05em}
\cmark~~~~~~~~~~ & 80.6 & 80.7 & 81.3 \\
\xmark~~(Ours) & 82.4 & 83.3 & ~83.0
\label{tab:dictionary_size}
\end{tabular}
}
\par}\smallskip

\noindent As training the forward-only encoder progresses faster (i.e., $m$ getting smaller), the degree of inconsistency of past samples with respect to a given query becomes larger. To observe the effect of the degree of inconsistency, experiments were conducted with various $m$. When past positive samples are included for training, using a larger $m$ to maintain consistency was crucial to yield a better accuracy. Nevertheless, using past positive samples expressed in the maximally consistent feature space (i.e., $m=0.999$ case) performed worse than the case where these past samples are excluded.

\subsection{Ablation Study: Similarity Measure.} 

As the proposed HE loss is defined using the similarity between the query sample and the key sample, we seek to find the most appropriate similarity measure by comparing the two methods: negative cosine similarity and the Euclidean distance. The similarity measures are defined as follows:

\begin{itemize}
    \item[$\bullet$] cosine: $d(x,y)=-\left(x/||x||_2\right)^T\left(y/||y||_2\right)$ 
    \item[$\bullet$] euclidean: $d(x,y)=||x-y||_2$
\end{itemize}

\noindent The table below shows mAP of models trained with the HE loss using different similarity measures:\smallskip

{\centering
\noindent
\setlength{\tabcolsep}{29.0pt}
\renewcommand{\arraystretch}{0.9}
{\small
\begin{tabular}{c|ccc}
\centering
\multirow{2}{*}{w/ Triplet loss} & \multicolumn{2}{c}{w/ HE loss} \\
& cosine & euclidean \\\specialrule{.15em}{.05em}{.05em}
79.4 & 82.9 & {\bf 83.3}
\label{tab:similarity}
\end{tabular}
}
\par}\smallskip

\noindent Regardless of the similarity measure, our model using the HE loss outperforms the Triplet loss. Although subtle, using the Euclidean distance yields the better than negative cosine similarity. Accordingly, we have used the Euclidean distance for the HE loss throughout all experiments.

\subsection{Ablation Study: Backbone Components}

As studied and suggested by~\cite{LHeArXiv2020}, we used a ResNet-50 architecture with new components as the backbone for optimal accuracy. The notable components are auto-augment pre-processing (AutoAug), instance batch normalization (IBN), and non-local modules (non-local). To verify that using the proposed HE loss is effective regardless of the influence of these components, we performed a set of ``Triplet loss vs. HE loss'' experiments where several variations of ResNet-50 were tested including the vanilla version as shown below: \smallskip

{\centering
\noindent
\setlength{\tabcolsep}{11.0pt}
\renewcommand{\arraystretch}{0.9}
{\small
\begin{tabular}{ccc|c|c}
\centering
AutoAug & IBN & non-local & w/ Triplet & w/ HE (gain)\\\specialrule{.15em}{.05em}{.05em}
\xmark & \xmark & \xmark & 77.7 & {\bf 80.2} \textcolor{teal}{(+2.5)} \\\hline
\xmark & \textcolor{red}{\cmark} & \textcolor{red}{\cmark} & 79.4 & {\bf 83.2} \textcolor{teal}{(+3.8)} \\
\textcolor{red}{\cmark} & \xmark & \textcolor{red}{\cmark} & 78.6 & {\bf 82.1} \textcolor{teal}{(+3.5)} \\
\textcolor{red}{\cmark} & \textcolor{red}{\cmark} & \xmark & 78.7 & {\bf 82.8} \textcolor{teal}{(+4.1)} \\\hline
\textcolor{red}{\cmark} & \textcolor{red}{\cmark} & \textcolor{red}{\cmark} & 79.7 & ~{\bf 83.3} \textcolor{teal}{(+3.6)}
\label{tab:vs_triplet_loss}
\end{tabular}
}
\par}\smallskip

\noindent Using the proposed HE loss consistently yields better accuracy than using the Triplet loss regardless of which components are being used.

%
%
\bibliographystyle{splncs04}
\bibliography{egbib}
\end{document}